# Towards automated mobile-phone-based plant pathology management

Nantheera Anantrasirichai, Sion Hannuna and Nishan Canagarajah

**Abstract**

This paper presents a framework which uses computer vision algorithms to standardise images and analyse them for identifying crop diseases automatically. The tools are created to bridge the information gap between farmers, advisory call centres and agricultural experts using the images of diseased/infected crop captured by mobile-phones. These images are generally sensitive to a number of factors including camera type and lighting. We therefore propose a technique for standardising the colour of plant images within the context of the advisory system. Subsequently, to aid the advisory process, the disease recognition process is automated using image processing in conjunction with machine learning techniques. We describe our proposed leaf extraction, affected area segmentation and disease classification techniques. The proposed disease recognition system is tested using six mango diseases and the results show over 80% accuracy. The final output of our system is a list of possible diseases with relevant management advice.

## 1. Introduction

More than half of the population in developing countries rely on farming for their livelihood. Whilst most are familiar with conventional farming practices, they are often ill positioned to promptly deal with diseases and plant infestations affecting their crops. Current advisory systems tend to be generic and are not tailored to specific plots or farms. This work forms part of a proposed agriculture disease mitigation system which aims to provide a mobile phone based 'Farmer to Expert' service, which facilitates prompt access to disease and pest mitigation advice. In order to enable the expert to provide relevant information, they are provided with data about the farmers' crops, including soil type and pH, crop variety and the pesticides and fertilizers being used.

The text-based information available to the expert will be augmented with mobile phone photos of the crops, which have been captured and uploaded by the farmer at the time of advice being sought. Unfortunately, the photos taken will be sensitive to a number of factors including camera type and lighting incident on the scene. Ideally, the images would be processed in such a way as to provide the expert with a visual representation of the affected crops that reflects the true nature of the scene. This is one aspect of the system that is specifically addressed in this paper.

In its current form, the system's diagnoses rely solely on the experts' assessments of processed images, farmers' profiles and personal correspondence with the farmers. We believe its efficacy could be improved by augmenting it with computer vision algorithms to automatically / semi-automatically identify diseased regions in these images and classify those regions according to disease type. It is anticipated that this system would assist, rather than replace the expert disseminating advice: a list of potential matches could be provided with corresponding likelihoods.

In this paper we present the tools that concern the protocols for standardising and analysing the mobile phone photographs, as well as the use of them for initially solving the disease problems and their managements. In terms of research contributions we offer i) *colour correction* to standardise image appearance, ii) *leaf extraction* to prepare for identifying a particular plant's species and assessing its health, and iii) *disease recognition* to find the possible disease affecting the crop. The rest of the paper demonstrates the application tools we developed together with the technical explanations and some results.

## 2. Tools

We have created a graphical user interface using Matlab GUIs (Mathworks, 2011) as illustrated in Figure 1. The interface allows users to perform colour correction, leaf extraction, disease marking and disease identification. For colour correction, a custom colour chart is selected. In section 3.1, we introduce a chart specifically tailored to plant foliage disease analysis. However, an alternative chart that best represents the plant species being analysed maybe more appropriate. For example, if the user is analysing bananas the chart should emphasise yellow coloured patches. Users' colour chart patch values may be automatically extracted from calibrated pictures or the user can input them manually. Users may also set some parameters to fine tune colour correction. Colour correction results may be viewed immediately and or saved.

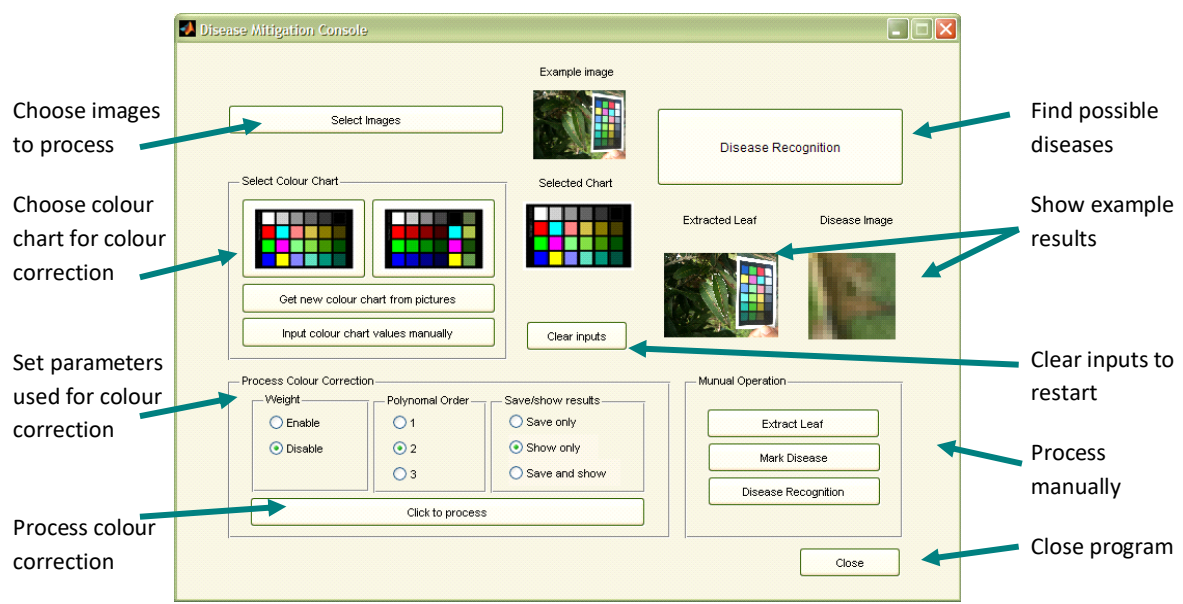

**Figure 1: User interface for agricultural advice**

To identify the disease on the leaf we provide one click process 'Disease Recognition' which includes extracting the target leaf, performing the automatic disease recognition and displaying the possible diseases with details and management method. The users are also able to operate these processes manually by choosing options in the manual operation section. The technical details of the leaf extraction and the disease recognition are described in section 4 and 5, respectively.

## 3. Colour Correction

The appearance of images depends on various elements. On the acquisition side, it is influenced by the interaction of the scene's content, the lighting conditions and the capture device. On the display side it is affected by the screen set-up and its local environment with both influencing how the human visual system perceives the scene (Fairchild, 2005). In this section we describe our custom colour chart and the use of it with the proposed colour mapping to standardise colour images.

### 3.1 Custom Colour Chart

We propose a cheap and easily implemented approximation to such a solution that makes use of automatically detected colour charts facilitating its integration into colour-managed imaging pipelines. The motivation for maximising how realistic the plant images are is that a great deal of information about the plants' health may be gleaned from their colours. In addition to striving for perceptual consistency, objective colour consistency increases the potential for fully exploiting colour as a feature in automated systems. The custom target shown in Figure 2 offers three groups of colour patches. Group 1 is composed of achromatic colours. Group 2 includes primary (RGB in 100% and 50% opacity) as well as secondary colours (CMY). The last group, 3, contains green tones created in CIE $L^*a^*b^*$ colour space with $L^*$ values of 25, 50 and 75 as well as $a^*$ and $b^*$ set to -65/65, -65/0 and 0/65. These colour charts are printed on professional matt colour papers so that different angles to the mobile phones will not cause reflection on the image.

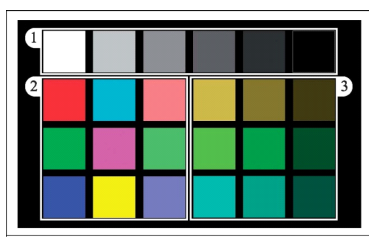

**Figure 2: Custom colour target designed to preferentially map plant colours and enable automatic**

### 3.2 Colour Chart Detection

Our custom colour chart displays concentric black and white borders. This arrangement facilitates a morphologic process and line detection. For automatic chart segmentation, the colour image is firstly converted into greyscale. The intensity image is then binarised such that the black and white borders of the chart are completely black white respectively. Next, the black and the white are reversed so that the colour patch areas on the colour chart which now appear in white are filled with black colour and an erosion process can be applied to clean the small areas. Then a Hough transform (Duda and Hart, 1972) is employed to detect the eroded black chart which is then used as the marker to reconstruct the original chart. Finally morphological opening is applied to find chart's corners and if four corners are found, 24 colours from colour chart are read.

### 3.3  Deducing a Transform Matrix

The aim of our system is to standardise images from low quality cameras such that they are both consistent with one another and provide a faithful representation of the scene they depict. The patch values for the source image to be processed and the ground truth image are used to deduce a transformation for the source image. This process is analogous to the characterisation of imaging devices including digital cameras. Various characterisation methods (Luo and Rhodes, 2001) essentially involve deducing a mapping for targets that have known device independent CIE XYZ values. Linear and polynomial transformations have been used to this end as well as neural networks. However, an important distinction for our application should be noted: normally characterisation is done once for a given device. Hence it is important that the mapping used generalises for all conceivable scenes and lighting conditions.

To exploit this stipulation we have experimented with weighting the relative cost for the residual error of particular patches under the deduced transformation according to how numerous image pixels closest to that patch are in a given image. Hence, in most of our images greens and browns would be favoured as they primarily depict plants. For the quadratic and linear transformations we utilise, this lends itself to a weighted least squares optimisation. The colour targets we tested only have 24 patches. The quadratic and linear transformations considered here produce perceptually convincing results. For more details the reader are referred to Hannuna et al. (2011b). The results of the proposed colour correction are shown in Figure 3. The same scene was captured from five different cameras. After applying colour mapping these images have similar colour tone. The result of the SamsungD500 shows significantly improvement which implies that the proposed scheme can make the image of the cheap phone more informative.

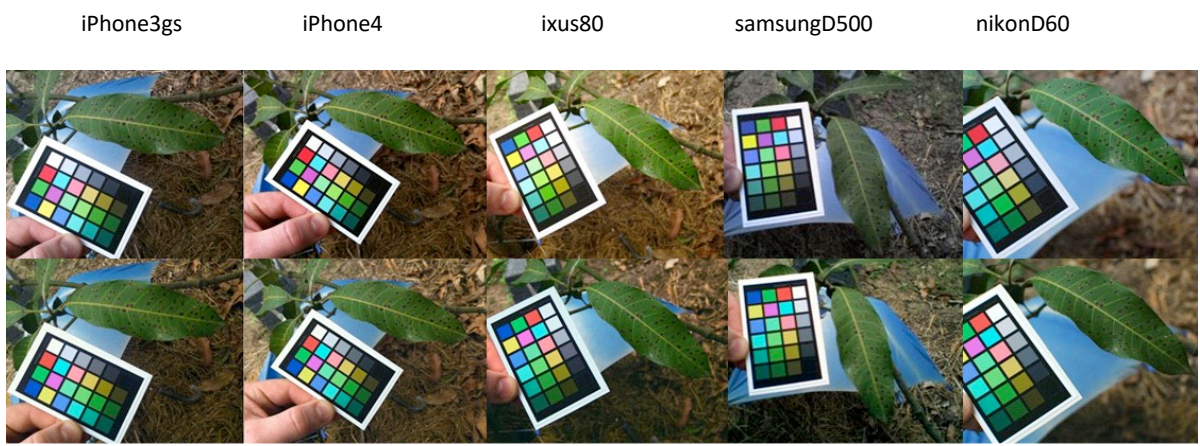

iPhone3gs     iPhone4     ixus80     samsungD500     nikonD60

**Figure 3:  Subjective results of colour correction. Above: original images taken from several cameras.**

## 4. Leaf Extraction

Plant foliage may be used to identify a particular plant's species and assess its health (Wand et at., 2003). Traditionally, these tasks would be carried out manually by an expert which is both time consuming and vulnerable to subjective variation. Automatic plant disease analysis using computers with/without supervising is nowadays widely considered. It, however, requires an image of a complete, isolated leaf as input to the process before further analysing other information in the leaf. This section describes the leaf extraction that we provide in our tools (Anantrasirichai et al., 2010).

The leaf extraction starts with the user roughly marking the target leaf using the green pen as shown in Figure 4. The user can also adjust the size of the pen to suit the target leaf size. Whilst the system is being told where the leaf is, the background marker is automatically created using colour, intensity and texture. The non-green areas are detected using the Otsu threshold (Otsu, 1979) and the high detail areas, such as grass, measured using a local entropy descriptor, are defined as the background marker. Subsequently the markers are used as local minima in the gradient magnitude map of the greyscale image calculated using a Sobel filter. With two markers, leaf and background, two regions are segmented with a watershed transform (Vincent and Soille,1991) and the region with leaf marker is selected as the extracted leaf.

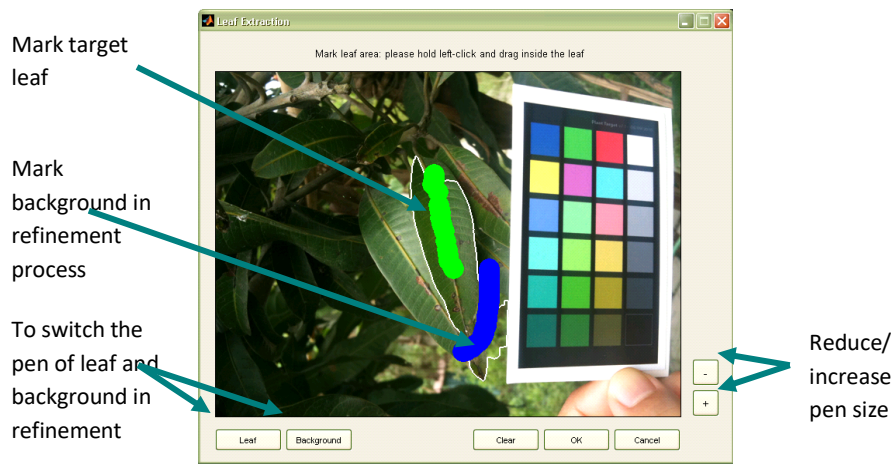

**Figure 4: User interface for leaf extraction**

In general, the watershed transform produces fast and acceptable results. In some cases, however, the gradients of the target leaf are insufficient for differentiating it from its background. We therefore provide a labelling tool where the user can annotate both the leaf and the problematic background region. The provided tool is demonstrated in Figure 4. Some results are shown in Figure 5. These results are used in the disease recognition process later.

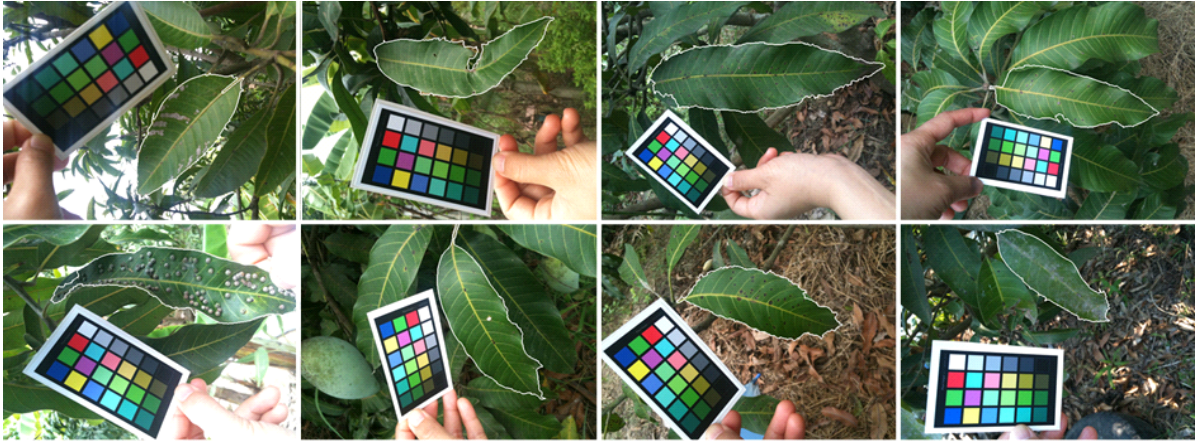

**Figure 5: Examples of extracted leaf results**

It is good to note here that the segmentation can be more accurate by drawing the outline of the leaf. However, it is not suitable for automating the leaf segmentation process. The idea behind marking only the middle of the leaf would help speeding the process. Say the user has 100 photos. He will be asked to just make a quick mark each image and leave the computer to automatic process of disease recognition.

## 5. Disease Recognition

Once the leaf has been extracted, the margin, shape and venation can be estimated in order to identify the plant's species, whilst the colours of the segmented leaves can be utilized to determine the plant's nutritional history and health. Machine learning techniques may be employed to classify different plant diseases. Several approaches have been used for this purpose, including: Decision Trees, K-means, artificial neural networks (ANNs) (Moshou et at., 2004), and support vector machines (SVM). Camargo and Smith (2009) reported a machine vision system for identifying plant diseases using the SVM. Several feature vectors were studied and showed that texture features improve the classification accuracy along to colour and shape features. Rumpf et al. (2010) reported that SVM's perform better than ANNs and Decision Trees for supervised prediction. During the classifier's training phase, human experts are required to facilitate supervised learning by assigning disease labels to a set of images comprising the training dataset. In this paper this expert-labelled dataset is comprised of six pathologies which manifest symptoms on mango leaves. These are: Anthracnose, Gall flies, Grey leaf spot, Red-rust, Powdery mildew and Sooty mould (TNAU Agritech Portal, 2011). The diseased leaf regions are extracted with the proposed technique in section 5.1 and are classified with the method in section 5.2. Note that the input images having the longer length more than 600 pixels are reduced the size to 600 pixels to decrease the computation time.

### 5.1 Affected Area Extraction

The areas affected by disease on the extracted leaves are isolated using the following conditions. Firstly, the affected areas are usually non-green and hence the histogram of the hue value may be thresholded to isolate the non-green areas. Secondly, the primary vein is detected and removed from the non-green area. This is necessary because the primary vein usually has a lighter or 'yellower' colour which could possibly be identified as the affected area by the automated system. By considering the orientation of a leaf's binary silhouette, it is possible the limit the angles searched whilst segmenting its primary vein. Third the white and black spots are always included in the affected areas since they are likely to be the disease areas.

The affected areas are extracted with the size range between 10x10 pixels to 25x25 pixels. The size of the affected area is limited in order to prevent generating areas which are too small and don't have enough useful information, or too big where the extracted features could be dominant. The affected areas which are larger than 25x25 pixels are labelled, as some diseases generally cover large areas of the leaf, e.g. Sooty mould and Powdery mildew. These labels will subsequently be used to reduce the possibility of the disease being recognised as one which only affects small leaf areas, e.g. Anthracnose and Gall flies. Figure 6 (left) illustrates the results of the proposed affected area detection. These detected areas will be used in the disease recognition where the output is the average of the results of all detected areas.

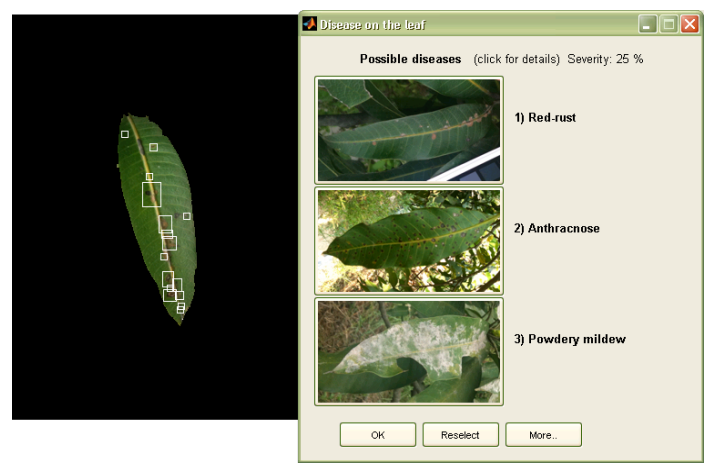

**Figure 6: Result of extracting disease areas automatically and list of possible disease obtained from disease recognition**

### 5.2  Disease Recognition

We employed a multi-class support vector machine (SVM) (Chang and Lin, 2002) to classify our test dataset. The classier was trained on six distinct diseases. The task of training and classification may be subdivided into three processes: i) computing feature vectors for training; ii) training the SVM classifier: this involves deducing a hyperplane (decision surface) that subdivides the data into (ideally) non-overlapping groups; and iii) predicting a label of the test dataset using the SVM model.

### 5.2.1  Feature Extraction

Colour and texture variation are the two main features appearing differently for each crop disease. For example, the Anthracnose produces black spots and small blisters with a yellow border, whilst the red-rust disease causes brown spots with a smoother surface texture. It is these feature variations which differentiate class labels for the SVM. For the colour features, the disease areas of which the colours are usually in RGB format are converted to HSV format. The average values of red, green, blue and hue are subsequently calculated from the disease affected areas. For the texture features we calculate contrast, correlation, energy and homogeneity from a grey level co-occurrence matrix (GLCM) (Haralick et at., 1973). We also find entropy to measure a randomness of the greyscale image. The mathematical formulations of the colour and texture features are tabulated in Table I. These candidate features are tested individually and in group. Some testing results of grouping features are shown in Table II. We can see that using all textures and colour features give best classification accuracy. The results of using individual feature are not shown in the table as they provide significant small accuracy (~ 50-60%). Note that if the images are noisy, the pre-processing denoising method should be applied. The simplest method is using mean filter which slightly deteriorate texture information. More complex methods give better results but high computation time. The reader is referred to the article of Antoni et at. (2004).

As the extracted affected areas obtained from 5.1 are rectangular but the actual disease areas are not, we mask the affected areas before computing the vector features with Equation 1.

$$M_{ij} = \begin{cases} 1 & (i-h/2)^2 + (j-w/2)^2 < (h/4+w/4)^2 \\ 0 & otherwise \end{cases}$$ 

Equation 1

where $M_{ij}$ is the mask at pixel position $i$ and $j$ of the affected area with the size of $h$ x $w$. The masked RGB and HSV images are converted to greyscale images and are used to find GLCMs separately. Each greyscale image creates four GLCMs from four orientation angles of 0, 45, 90 and 135 degree with the distance of 1 between the pixel of interest and its neighbour. Therefore 4 colour features and 120 texture features are generated in total. Finally the feature vectors are scaled to the range [-1,1] so that the greater numeric values will not dominate the smaller numeric ones. The scaling parameters will also be applied to the new data in the recognition process

**Table I. Formulas of the colour and texture features**

| Feature | Equation |
|---|---|
| Average Colour | $X = \left(\sum_{ij} M_{ij} x_{ij}\right) / \left(\sum_{ij} M_{ij}\right)$, $x = red, green, blue, hue$ |
| Contrast | $Contrast = \sum_{ij} |i-j|^2 g_{ij}$, $g_{ij} = $ GLCM $at\ (i,j)$ |
| Correlation | $Correlation = \sum_{ij}(i-\mu_i)(j-\mu_j)g_{ij}/(\sigma_i \sigma_j)$, $g_{ij} = $ GLCM $at\ (i,j)$ |
|  | $\mu_i = \sum_i i g_{ij}$, $\mu_j = \sum_j j g_{ij}$, $\sigma_i = \sqrt{\sum_i (i-\mu_i)^2 g_{ij}}$, $\sigma_j = \sqrt{\sum_j (j-\mu_j)^2 g_{ij}}$ |
| Energy | $Energy = \sum_{ij} g_{ij}^2$, $g_{ij} = $ GLCM $at\ (i,j)$ |
| Homogeneity | $Homogeneity = \sum_{ij} g_{ij}/(1+|i-j|)$, $g_{ij} = $ GLCM $at\ (i,j)$ |
| Entropy | $Entropy = -\sum_{ij} g_{ij} \log(g_{ij})$ |

### 5.2.2 Multi-class Support Vector Machine

We implemented the multi-class SVM using the LIBSVM – A Library for Support Vector Machines (Chang and Lin, 2011). The radial basis function (RBF) kernel is used because it allows the decision surfaces to be determined for data which is not linearly separable. Cross-validation is also used to find the best parameters for RBF kernel so that the classifier can accurately predict unknown data and can prevent the over fitting problem.

**Table II. Results of selecting a subset of relevant features**

| Feature group | Classification accuracy % |
|---|---|
| RGB + Contrast | 74.26 |
| RGB + Correlation | 70.42 |
| RGB + Energy | 72.08 |
| RGB + Homogeneity | 72.98 |
| RGB + Entropy | 71.35 |
| HSV + Contrast | 73.21 |
| HSV + Correlation | 70.04 |
| HSV + Energy | 71.47 |
| HSV + Homogeneity | 72.52 |
| HSV + Entropy | 70.15 |
| R + all textures | 88.47 |
| G + all textures | 87.25 |
| B + all textures | 92.66 |
| RGB + all textures | 92.15 |
| H + all textures | 90.24 |
| S + all textures | 88.84 |
| V + all textures | 89.04 |
| HSV + all textures | 90.11 |
| RGB + HSV + all textures | 92.65 |

In this paper, 120 labelled disease images of each mango disease are used to train the SVM classifier. When classifying previously unseen disease images, its features are extracted (section 5.2.1), scaled and applied to the SVM predictor. The predictor yields a probability for each of the six diseases it was trained on. For each of the pathological areas in a given example, the user is provided with a sorted list of diseases types with their associated probabilities. We also allow the users to select the affected area manually. As shown in Figure 6 the 'Reselect' button is linked to the manual selection function. The leaf image will be popped up and the user is asked to click at the top-right and bottom-left corners of the target affected area. The manually selected affected area is processed with the SVM predictor and once again an ordered list is provided.

### 5.2.3 Disease symptoms and management display

Figure 6 (right) shows the list of possible diseases. Three diseases are displayed each time. The users can click to see more diseases or go back to the more probable diseases in the list. A picture of each disease is provided so that the users can compare it with their own photo. When the picture is clicked, the details of the symptoms and the way to manage this disease are displayed as shown in Figure 7. If the automatic affected area extraction is used, the approximate severity is also calculated from the total affected area divided by the leaf area. This parameter provides a qualitative measure of severity of the disease's condition.

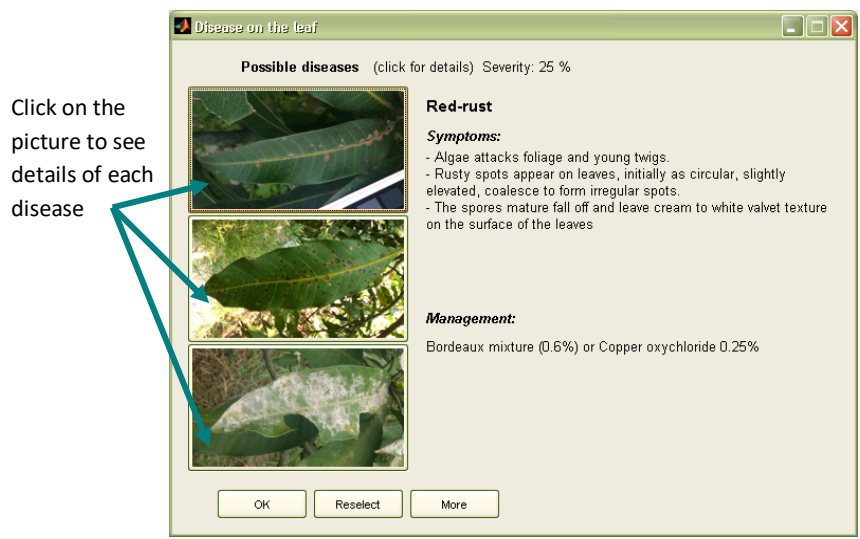

Figure 7: Details and management methods of each plant diseases

### 5.2.4 Results

We evaluated our proposed system with two data sets. The first set includes 600 disease images selected manually which are comprised of 100 images of each disease. The results shown in Table III reveal that the proposed disease recognition achieves 89.83% accuracy. Furthermore, second and third ranked predictions account 7.67% and 2.33% diseases respectively. The second set we tested contains 150 leaf images (25 images for each disease). This test included the automatic affected area extraction and disease recognition.

**Table III. Number of each disease extracted manually appearing in each order of the possible disease list**

| Disease | #1 | #2 | #3 | #4 | #5 | #6 | total |
|---|---|---|---|---|---|---|---|
| Anthracnose | 92 | 7 | 1 | - | - | - | 100 |
| Gall flies | 89 | 8 | 3 | - | - | - | 100 |
| Grey leaf spot | 91 | 8 | 1 | - | - | - | 100 |
| Powdery mildew | 90 | 5 | 4 | 1 | - | - | 100 |
| Red-rust | 90 | 8 | 2 | - | - | - | 100 |
| Sooty mould | 87 | 10 | 3 | - | - | - | 100 |
| total | 539 | 46 | 14 | 1 | 0 | 0 | 600 |
| % | 89.83 | 7.67 | 2.33 | 0.17 | 0 | 0 | 100% |

**Table IV. Number of each disease leaf appearing in each order of the possible disease list**

| Disease | #1 | #2 | #3 | #4 | #5 | #6 | total |
|---|---|---|---|---|---|---|---|
| Anthracnose | 22 | 2 | 1 | - | - | - | 25 |
| Gall flies | 20 | 4 | 1 | - | - | - | 25 |
| Grey leaf spot | 18 | 5 | 2 | - | - | - | 25 |
| Powdery mildew | 20 | 4 | - | 1 | - | - | 25 |
| Red-rust | 22 | 2 | 1 | - | - | - | 25 |
| Sooty mould | 17 | 5 | 2 | - | 1 | - | 25 |
| total | 119 | 22 | 7 | 1 | 1 | 0 | 150 |
| % | 79.33 | 14.67 | 4.67 | 0.67 | 0.67 | 0 | 100% |

Table IV shows that the automatic affected area decreases the accuracy of the recognition to 79.33% which is approximately 10% reduction. However, the second most probable disease accounts for a further 14.67% giving comparable accuracy for the automatic and manual approaches, when the top two are considered together (remember the user is provided with images of the diseases to make up their own mind).

## 6. Conclusions

In this paper we present an application for semi-automatic disease recognition for use in conjunction with a mobile-phone disease mitigation system. The first contribution demonstrates the efficacy of our custom colour chart and the use of a weighted least squares formulation in standardising images for pathology diagnosis. The second contribution is comprised of leaf extraction, affected area segmentation and disease classification and recognition. A marker-controlled watershed segmentation is used for leaf extraction. The affected areas are automatically identified using hue and gradient information. For the disease classification a support vector machine (SVM) is employed. The predicted results are ranked according to their probabilities, and then their symptoms and management are displayed to the user. Here six mango diseases are used for testing the framework. We found that the proposed disease recognition achieves about 90% and 80% accuracy when the affected areas are selected manually and automatically respectively.